\documentclass[sigconf,balance=false]{acmart}

\setcopyright{none}
\acmConference{}{}{}{}
\acmBooktitle{}
\acmYear{}
\copyrightyear{}
\acmDOI{}
\acmISBN{}
\renewcommand{\footnotetextcopyrightpermission}[1]{}
\AtBeginDocument{\fancyhead[LE,RO]{}}

\usepackage{algorithm}
\usepackage{algpseudocode}
\usepackage{colortbl}
\usepackage{balance}
\usepackage{stfloats}
\definecolor{GuardianTint}{RGB}{232,241,251}
\definecolor{RankFirst}{RGB}{220,239,222}
\definecolor{RankSecond}{RGB}{255,243,204}
\definecolor{RankThird}{RGB}{211, 211, 211}
\newcommand{\rankfirst}[1]{\cellcolor{RankFirst}\textbf{#1}}
\newcommand{\ranksecond}[1]{\cellcolor{RankSecond}\textit{#1}}
\newcommand{\rankthird}[1]{\cellcolor{RankThird}#1}

\newcommand{\sd}[1]{\textsubscript{$\pm$#1}}

\newcommand{\sys}{GuardianAgentFlow}

\title[GuardianAgentFlow: Multi-Agent Web Privacy Protection]{GuardianAgentFlow: Multi-Agent Coordination\texorpdfstring{\\}{ }for Web Privacy Protection}

\author{\texorpdfstring{%
  Ruiyi Yang\quad Gayathri Lihinikaduarachchi\quad Rahat Masood\\[2pt]
  Flora D. Salim\quad Salil S. Kanhere
}{Ruiyi Yang; Gayathri Lihinikaduarachchi; Rahat Masood; Flora D. Salim; Salil S. Kanhere}}
\affiliation[obeypunctuation=true]{%
  \department{School of Computer Science and Engineering}\\
  \institution{UNSW} \city{Sydney}, \country{Australia}\\[2pt]
  \texttt{\{ruiyi.yang, g.lihinikadu\_arachchillage\}@student.unsw.edu.au}\\
  \texttt{\{rahat.masood, flora.salim, salil.kanhere\}@unsw.edu.au}
}
\renewcommand{\shortauthors}{Yang et al.}
\renewcommand{\authors}{Ruiyi Yang\and Gayathri Lihinikaduarachchi\and Rahat Masood\and Flora D. Salim\and Salil S. Kanhere}


\begin{document}
\begin{abstract}
Web agents need to use private context without unnecessarily disclosing it. Effective protection must distinguish between requests that may be sent after sanitization and those that must be withheld, while preserving the information needed for authorized tasks. We present \sys{}, a multi-agent framework that mediates outgoing Web requests through an explicit release contract. A dedicated restriction check grounds permission in the user's instruction; a shared protection scope aligns rewriting with disclosure checking; and candidate-grounded feedback guides revision before transmission. Experiments on Web requests with human-reviewed intent labels show improved request-contract success, while controlled comparisons and role ablations identify the contributions of restriction handling, policy evidence, and candidate checking. Evaluations of action proposals from PrivacyLens and AgentDAM demonstrate complementary protection when mediation is composed with privacy-aware agents. Experiments with smaller mediation models show that the coordination benefit persists with reduced model capacity. Dedicated-server measurements show lower mediation latency than full-policy reasoning, while real-site browser tests quantify integration overhead. Code is available at https://anonymous.4open.science/r/guardianAgentFlow/
\end{abstract}
\maketitle

\section{Introduction}
\label{sec:introduction}

Web agents increasingly act on information that combines a legitimate task with details the recipient does not need. A customer-service request may include a billing question, a private contact number, and an instruction about what may be sent. ``Send the question without my number'' authorizes a sanitized request that retains the billing question. ``Do not send this request'' requires withholding the entire payload. Effective privacy protection must therefore determine whether transmission is authorized and preserve the information needed to complete authorized tasks.

Meeting these obligations requires coordination at the request boundary. Policy analysis characterizes the destination's stated practices~\cite{harkous2018polisis,cui2023poligraph}, and retrieval supplies relevant evidence~\cite{lewis2020rag}. Contextual privacy agents govern what an acting agent sees or discloses~\cite{bagdasarian2024airgap,wang2025privacyinaction}, while policy-conditioned anonymization adapts sensitive content to privacy requirements~\cite{yang2026guardianagent}. These judgments jointly determine the outgoing action: user permissions constrain transmission, protected and task-required information guide rewriting, and candidate verification governs release. Disagreement across these stages can cause unintended disclosure, excessive removal of useful information, or rejection of an acceptable candidate.

We present \sys{}, a multi-agent framework that mediates outgoing Web requests through an explicit release contract. A dedicated restriction check extracts user constraints together with supporting spans from the instruction. At runtime, it combines these constraints with the Policy Agent's decision, based on destination evidence, and establishes a shared scope of protected and task-required information. The Anonymization Agent and Judger use this scope for rewriting and verification, with disclosure findings grounded in quoted spans of the outgoing candidate. A coordinator manages bounded revision and release, while the Monitor binds the final decision to the intercepted request. This contract connects permission, transformation, and verification through explicit intermediate decisions that guide both execution and diagnosis.

Our evaluation examines request handling, the contributions of individual mechanisms, and execution cost. Constructed requests paired with public website policies measure useful minimization and requested withholding under human-reviewed intent labels. A newly generated request cohort tests the frozen configuration on different wording and private values. Action proposals from PrivacyLens~\cite{shao2024privacylens} and AgentDAM~\cite{zharmagambetov2025agentdam} evaluate protection across contextual-privacy and strict-minimization settings. Controlled comparisons and role ablations examine the contributions of restriction handling, policy evidence, shared scope, and candidate checking. Smaller-model experiments assess the design across model capacities, while dedicated-server and browser measurements quantify mediation performance and integration overhead.

The results show that \sys{} improves useful transformation and requested withholding over policy-based controls and a content-matched one-call mediator. On public action proposals, the shared protection scope helps preserve useful content, and composing mediation with privacy-aware agents provides additional protection. The coordination benefit persists with smaller mediation models. Compared with full-policy reasoning, \sys{} achieves higher request success at lower token cost and measured mediation latency; branch-specific timing links execution cost to the roles invoked for each request. We make three contributions:
\begin{itemize}
    \item A request mediation contract that composes user restrictions with destination-policy decisions and carries a shared protection scope through rewriting, checking, and release.
    \item A coordinated rewriting and verification procedure that uses candidate-grounded findings, bounded revision, and final eligibility checks to repair residual disclosures and control outgoing content.
    \item A system implementation and evaluation spanning human-reviewed requests, public action benchmarks, role and model-size interventions, dedicated-server performance under sequential and concurrent load, and real-site browser integration.
\end{itemize}
\section{Related Work}
\label{sec:related-work}

\paragraph{Policy understanding and evidence retrieval.}
Privacy-policy analysis connects natural-language commitments to website practices. OPP-115 provides annotations of policy practices, and PrivaSeer supports analysis across a large collection of Web policies~\cite{wilson2016policycorpus,srinath2021privaseer}. PrivacyQA and PolicyQA formulate question answering over policy text, and Polisis enables structured analysis and interactive querying~\cite{ravichander2019privacyqa,ahmad2020policyqa,harkous2018polisis}. PolicyLint detects contradictions within policies; PoliCheck and PoliGraph relate policy statements, entities, and observed information flows~\cite{andow2019policylint,andow2020policheck,cui2023poligraph}.

Retrieval supplies relevant evidence within a bounded context. BM25 provides lexical matching, and Sentence-BERT and dense passage retrieval support semantic matching~\cite{robertson2009bm25,reimers2019sbert,karpukhin2020dpr}. BEIR and Lost in the Middle highlight the importance of retrieval conditions and evidence placement~\cite{thakur2021beir,liu2024lostmiddle}. Retrieval-augmented generation conditions outputs on retrieved documents~\cite{lewis2020rag}. Subsequent methods adapt evidence acquisition and assessment: FLARE retrieves during generation, Adaptive-RAG selects retrieval strategies according to question complexity, and Self-RAG and CRAG assess retrieval needs or the quality of the retrieved support~\cite{jiang2023flare,jeong2024adaptiverag,asai2024selfrag,yan2024crag}.

\paragraph{Contextual privacy and multi-agent coordination.}
Contextual integrity characterizes privacy through the participants, information, and transmission conditions of an information flow~\cite{nissenbaum2004contextual}. CONFAIDE evaluates contextual privacy judgments, and PrivacyLens examines the relationship between stated judgments and disclosures in agent actions~\cite{mireshghallah2024confaide,shao2024privacylens}. AirGapAgent separates task-based minimization from third-party conversation, and PrivacyChecker extracts information flows and assesses their appropriateness during agent execution~\cite{bagdasarian2024airgap,wang2025privacyinaction}. MINIM sanitizes local interface observations before remote processing, and the Firewalls framework separates input, data, and trajectory protection in dynamic agent networks~\cite{yu2026minim,abdelnabi2026firewalls}. These approaches introduce protection at different stages of an agent's information flow.

Recent work also examines cooperation among privacy roles. 1-2-3 Check decomposes contextual privacy reasoning into extraction and classification with iterative validation~\cite{li2025check}. PrivAct incorporates contextual privacy preferences into agents through multi-agent preference training~\cite{cheng2026privact}. PAC-Bench evaluates collaboration under privacy constraints and identifies coordination failures involving premature disclosure, excessive abstraction, and privacy-induced hallucination~\cite{park2026pacbench}.

\paragraph{Runtime enforcement and untrusted content.}
Runtime protection connects policy requirements to constraints on generation and execution. LMQL constrains generation through a programming interface~\cite{beurerkellner2023lmql}. GuardAgent translates requirements into executable checks, Progent enforces privileges over tools and arguments, and CaMeL separates trusted control from untrusted data through capabilities and provenance~\cite{xiang2025guardagent,shi2026progent,debenedetti2025camel}. Research on indirect prompt injection identifies attacks carried through external content~\cite{greshake2023injection}, and StruQ strengthens the instruction--data boundary through structured queries and training~\cite{chen2025struq}. Classical protection principles provide a foundation for mediating access to protected resources~\cite{saltzer1975protection}.

\paragraph{Anonymization and candidate verification.}
Text anonymization balances disclosure reduction with the preservation of meaning~\cite{lison2021anonymisation}. TAB, SynthPAI, and PII-Masking provide complementary resources for studying annotated sensitive spans, personal-attribute inference, and identifier masking~\cite{pilan2022tab,yukhymenko2024synthpai,ai4privacy2024pii}. Work on personal-attribute inference motivates the protection of information conveyed through contextual clues as well as explicit identifiers~\cite{staab2024inference}. Adversarial anonymization, feedback-trained rewriting, and entity hiding and restoration provide different mechanisms for transforming sensitive content~\cite{staab2025anonymizers,shao2025agentstealth,chen2023has}. Policy-conditioned risk adaptation and verified escalation further connect transformation strength to privacy requirements and verification feedback~\cite{yang2026guardianagent}.

\paragraph{Evaluating privacy in agent actions.}
ToolEmu, AgentDojo, and InjecAgent evaluate tool-related risks and adversarial outcomes~\cite{ruan2024toolemu,debenedetti2024agentdojo,zhan2024injecagent}. AgentDAM measures disclosure during autonomous Web tasks~\cite{zharmagambetov2025agentdam}. PrivaCI-Bench evaluates legal and policy compliance, and CI-Work studies essential and sensitive information in enterprise workflows~\cite{li2025privacibench,fu2026ciwork}. These benchmarks motivate joint assessment of useful task execution and inappropriate disclosure. Research on model-based evaluation also examines agreement with human judgments and variability across evaluators~\cite{zheng2023llmjudge}.

Building on this work, \sys{} coordinates privacy decisions at the outgoing-request boundary through an explicit release contract. The contract composes user restrictions with destination-policy decisions, supplies rewriting and checking with shared protected and task-required information, and grounds disclosure findings in the candidate to be transmitted. Bounded revision and request-bound release connect these judgments to execution, supporting both useful minimization and requested withholding. This interface also enables composition with upstream privacy-aware agents.
\section{System Design}
\label{sec:method}
\begin{figure*}[t]
 \centering
 \includegraphics[width=\textwidth]{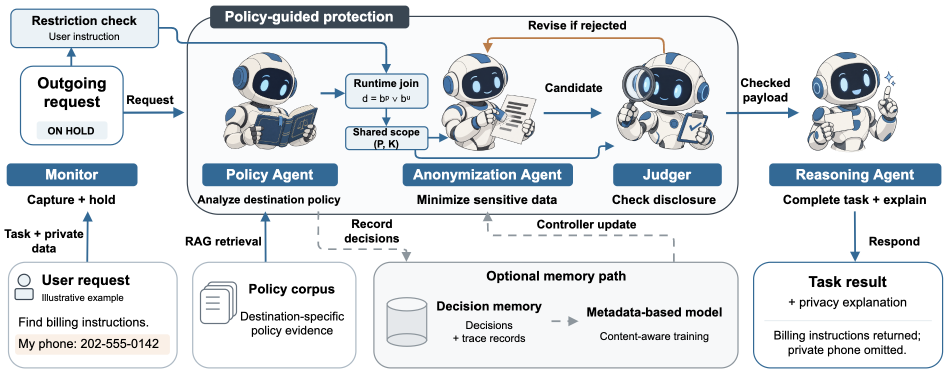}
 \caption{\sys{} and its information flow. Blue denotes online mediation; orange denotes revision feedback. The runtime composes the Policy Agent's evidence-based decision with separately extracted user restrictions. Rewriting and checking share protected and required information. The optional memory and training path is disabled in the evaluated configuration.}
 \label{fig:system-architecture}
 \Description{The Monitor holds an outgoing request. The runtime composes policy decisions with extracted user restrictions to determine permission and protection scope. The Anonymization Agent and Judger share this scope for rewriting and checking before release to task reasoning. The optional memory path is separate.}
\end{figure*}

\sys{} coordinates five components through an explicit request contract: the Monitor holds the outgoing request. The Policy Agent proposes a policy-based action and protection scope, which the runtime composes with separately extracted user restrictions. The Anonymization Agent proposes a replacement, the Judger checks it, and the Reasoning Agent uses permitted information for task completion and explanation (Figure~\ref{fig:system-architecture}). The roles can share model weights; a deterministic runtime composes their outputs and controls release.

\subsection{Request State and Mediation Contract}
\label{sec:system-model}
In \sys{}, the request contract specifies transmission permission,
protected and task-required information, and the conditions under which the runtime may release an outgoing payload. The Monitor $M$ maps an intercepted request to the event
\begin{equation}
 e=(\iota,m,x,u,t,c,\eta),
 \label{eq:event}
\end{equation}
where $\iota$ identifies the request, $m$ contains action and destination metadata, $x$ is its payload, $u$ is the user's instruction, $t$ is the task, and $c$ is additional context. The source flag $\eta\in\{\mathsf{user},\mathsf{agent}\}$ distinguishes a user-authored outgoing request from an action proposed by an agent. This distinction determines how the system maps an omission instruction to payload content.

The Monitor retains $x$ while mediation produces $(d,\Omega,y,o)$: decision $d\in\mathcal D=\{\mathsf{allow},\mathsf{transform},\mathsf{deny}\}$, protection scope $\Omega$, eligible outgoing payload $y$, and terminal outcome $o$. We use $y=\bot$ for no outgoing payload. The terminal outcome $o$ distinguishes intentional denial from blocking when mediation cannot produce an eligible payload. Website evidence and proposed content are untrusted inputs; the local mediation service and its state form the trusted boundary. Human labels and benchmark reference facts belong to evaluation, outside $e$.

\subsection{Policy Evidence and Permission Composition}
\label{sec:system-grounding}
Let $E(e)$ be the destination-policy passages retrieved for the event using lexical and structured ranking. The Policy Agent $\Pi$ maps the event and this evidence to
\begin{equation}
 (b^p,\rho,P^p,K^p,z)=\Pi(e,E(e)),
 \label{eq:policy-contract}
\end{equation}
where $b^p$ is its proposed action, $\rho$ is risk, $P^p$ lists information to protect, $K^p$ lists task-required values, and $z$ explains the decision. The two lists translate contextual policy reasoning into an explicit transformation objective.

\begin{figure*}[t]
 \centering
 \includegraphics[width=\textwidth]{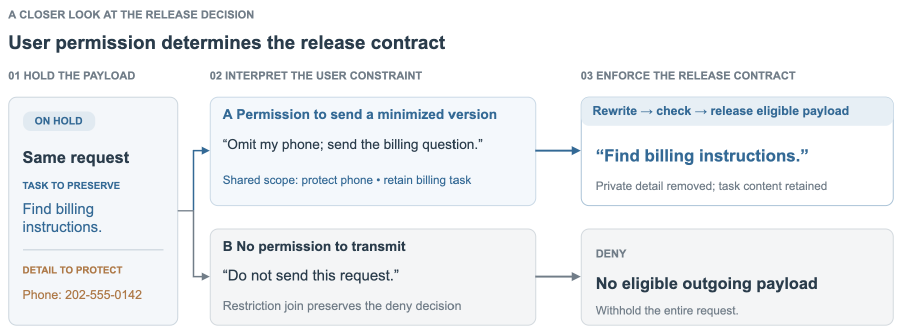}
 \caption{Two release contracts, illustrated with the same task and private detail. A content constraint permits a minimized request after rewriting and checking; a transmission constraint requires withholding even if the payload could be sanitized.}
 \label{fig:release-contracts}
 \Description{A held request contains a billing task and a private phone number. If the instruction permits sending without the phone, a shared protection list guides rewriting and checking before release. If the instruction forbids the whole request, the restriction join produces no eligible outgoing payload.}
\end{figure*}

A separate extractor $U(u)$ returns a restriction, its quoted source, and grounded omission items $O$. It reads the user instruction separately from destination evidence. The runtime validates the quote and supplements model extraction with explicit refusal rules. Write $R(u)=1$ for a grounded whole-request refusal and $G(u,x)=1$ for a grounded omission that names content in a user-authored outgoing request. The restriction action is
\begin{equation}
 b^u=\begin{cases}
 \mathsf{deny},&R(u)=1,\\
 \mathsf{transform},&\eta=\mathsf{user},\ G(u,x)=1,\\
 \mathsf{allow},&\text{otherwise}.
 \end{cases}
 \label{eq:user-restriction}
\end{equation}
Thus ``do not send this request'' selects denial, whereas ``send it without my number'' permits transformation (Figure~\ref{fig:release-contracts}). The Policy Agent interprets general privacy preferences in the context of the task and destination evidence.

Composition takes stronger action in an ordered decision space:
\begin{equation}
 d=b^p\vee b^u,\qquad
 \mathsf{allow}\preceq\mathsf{transform}\preceq\mathsf{deny}.
 \label{eq:restriction-join}
\end{equation}
The composed decision satisfies $d\succeq b^u$, preserving the action selected by the grounded restriction check. On the transform branch, let $O_e$ contain the grounded user omissions, or be empty when none apply. The shared scope is
\begin{equation}
 \begin{aligned}
 P&=P^p\cup O_e,\\
 K&=K^p\setminus\{k:\exists o\in O_e,\ o\sqsubseteq k\},\qquad
 \Omega=(P,K),
 \end{aligned}
 \label{eq:shared-scope}
\end{equation}
where $o\sqsubseteq k$ denotes a case-insensitive substring match. Explicit omission therefore takes priority over retaining a conflicting value. For an agent proposal ($\eta=\mathsf{agent}$), the Policy Agent derives $P^p$ and $K^p$ jointly from the user's instruction, task requirements, and private context; $O_e$ is empty. Grounded whole-request refusals enter the restriction join for both sources.

\subsection{Scope-Guided Rewriting and Grounded Checking}
\label{sec:system-transformation}
For $d=\mathsf{transform}$, the Anonymization Agent $A$ receives the original payload and the shared scope. At round $j$ it produces
\begin{equation}
 x^{(j)}=A(x;P,K,\ell_j,F_{j-1}),\qquad F_0=\varnothing,
 \label{eq:rewrite}
\end{equation}
where $\ell_j$ is rewriting strength and $F_{j-1}$ contains verified findings from the previous candidate. The objective is to remove $P$ while retaining $K$. Risk initializes the strength, which can escalate within a budget
of three rewrite/check rounds.

The Judger returns an overall confidence, candidate findings, and an error indicator:
\begin{equation}
 (\widetilde{s}_j,\widetilde{F}_j,\epsilon_j)
 =J(x^{(j)};P,K,c_{\mathrm{pub}}).
 \label{eq:judger}
\end{equation}
Here $\widetilde{s}_j$ is the reported \texttt{max\_confidence},
$c_{\mathrm{pub}}$ is public context such as the destination,
and $\epsilon_j=1$ marks a malformed or failed check.
A finding $f=(p,q,\gamma)\in\widetilde{F}_j$ identifies a protected
item $p$, a quoted candidate span $q$, and confidence $\gamma$.
The runtime forms $F_j$ by retaining findings whose quoted spans
occur in $x^{(j)}$ and match the protected scope lexically or
semantically; task-required values are exempted.
With an empty protection list, the checker uses the
original-content verification fallback.

The decision score combines verified finding confidences with
the Judger's overall confidence:
\begin{equation}
 s_j=
 \begin{cases}
  1, & \epsilon_j=1,\\
  0, & \epsilon_j=0,\ F_j=\varnothing,\\
  \max\{\widetilde{s}_j,\max_{f\in F_j}\gamma_f\},
     & \text{otherwise},
 \end{cases}
 \label{eq:verified-score}
\end{equation}
where $\gamma_f$ is the confidence of finding $f$.
The evaluated configuration uses $\tau=0.6$.
Verified findings guide subsequent rewrites within the revision
budget. A score below $\tau$ ends model revision, while failed
checks or unresolved disclosure prevent release.

The runtime can additionally delete grounded residual phrases when revision is exhausted, or when a checked candidate retains a protected value verbatim. This repair respects required values and is followed by a final Judger check and runtime validation. Let $x^\star$ be the resulting candidate, with verified score $s_\star$ and check status $\epsilon_\star$, and let $V(x,x^\star,\Omega)$ denote the adapter's structural and residual-content checks. Eligibility requires
\begin{equation}
 C(x^\star)=
 [\epsilon_\star=0]\land[s_\star<\tau]\land[V(x,x^\star,\Omega)=1].
 \label{eq:candidate-eligibility}
\end{equation}
The post-check detects protected strings and residual identifiers, with required values exempted where applicable. Task-required content guides rewriting and repair, and the evaluation assesses its retention.

\subsection{Release, Execution, and Traceability}
\label{sec:coordination-protocol}
\label{sec:system-release}
After validating the decision response, the runtime selects the outgoing payload:
\begin{equation}
 y(e)=\begin{cases}
 x,&d=\mathsf{allow},\\
 x^\star,&d=\mathsf{transform}\ \land\ C(x^\star),\\
 \bot,&\text{otherwise}.
 \end{cases}
 \label{eq:release}
\end{equation}
An invalid decision also yields $\bot$. The terminal outcome is $\mathsf{released}$ when an eligible payload is available, $\mathsf{denied}$ when the composed decision prohibits transmission, and $\mathsf{blocked}$ when mediation fails to produce an eligible payload, including invalid responses, failed checks, or unresolved disclosure. These outcomes support separate evaluation of requested withholding and completion of authorized tasks.

The contract imposes three coordination properties. First, Eq.~\ref{eq:restriction-join} preserves recognized user restrictions. Second, both $A$ and $J$ receive the same $\Omega$, so checking uses the protection objective supplied to rewriting. Third, Eq.~\ref{eq:candidate-eligibility} applies to the final transformed candidate, including any residual-deletion repair. These properties connect role outputs to release and make role replacement and ablation possible while retaining the surrounding interfaces.

The browser adapter binds $(\iota,y,o)$ to the held event and replaces its content only when $y\ne\bot$. For structured actions, content-bearing fields can change while routing fields retain the selected recipient. For released content, the Reasoning Agent computes $H(t,y,z)=(a_{\mathrm{task}},a_{\mathrm{privacy}})$, combining task completion with a privacy explanation. A trace records the event identifier, restriction source, $d$, $\Omega$, candidate checks, and terminal outcome, locating errors in permission, scope selection, rewriting, or transport.

\subsection{Conditional Computation and Deployment}
\label{sec:system-cost}
The five components are logical responsibilities, so deployment can use a common model server across reasoning roles. The Monitor and release coordinator remain deterministic. Policy interpretation and restriction extraction establish the action branch; rewriting and disclosure checking are invoked only for a transformation. This separates the cost of deciding permission from the additional cost of repairing content.

Let $T_M$ cover interception and local release processing, $T_E$ evidence retrieval, and $T_\Pi,T_U$ the policy and restriction calls. For a transformed request, let $T_{\mathrm{rev}}$ include the invoked rewrite/check rounds and any final residual repair. For sequential execution of the mediation stages, the latency is
\begin{equation}
 T(e)=T_M+T_E+T_\Pi+T_U
 +\mathbf{1}[d=\mathsf{transform}]T_{\mathrm{rev}}.
 \label{eq:mediation-cost}
\end{equation}
The revision budget bounds the number of rewrite/check rounds; early acceptance terminates the loop as soon as the candidate meets the release criterion. A permitted unchanged request or an intentional denial therefore avoids the rewriting loop entirely. Downstream task reasoning starts after release and is separate from this mediation cost. This decomposition motivates measuring latency by action branch as well as overall throughput.

\begin{table*}[!b]
 \centering
 \caption{Web-request effectiveness (\%). Transform/withhold: success by human intent; contract: correct release, lexical privacy, and required-phrase retention; semantic: adds Llama-70B disclosure checks. Original: 579 labeled requests; fresh: 278 new inputs. All methods achieve 100\% on allow in all runs. Every cell: three-run mean with sample SD in subscript. Tokens include all model roles.}
 \label{tab:native-comparison}
 \small
 \begin{tabular}{@{}lrrrrrr@{}}
  \toprule
  Method & Transform$\uparrow$ & Withhold$\uparrow$ & Contract$\uparrow$ & Semantic$\uparrow$ & Fresh$\uparrow$ & Tokens/task$\downarrow$\\
  \midrule
  Local rules & 0.0\sd{0.0} & 8.0\sd{0.0} & 36.5\sd{0.0} & 36.1\sd{0.0} & 53.6\sd{0.0} & \rankfirst{0\sd{0}}\\
  Full policy & 30.8\sd{0.2} & 91.7\sd{1.0} & 66.8\sd{0.3} & 66.3\sd{0.3} & 82.4\sd{0.6} & 8,625\sd{13}\\
  BM25-RAG & 22.7\sd{0.8} & \ranksecond{96.5\sd{0.0}} & 64.0\sd{0.3} & 63.2\sd{0.3} & 81.3\sd{0.6} & \ranksecond{1,339\sd{11}}\\
  \sys{} & \rankfirst{92.8\sd{0.0}} & \rankfirst{99.1\sd{0.0}} & \rankfirst{96.5\sd{0.0}} & \rankfirst{96.1\sd{0.1}} & \rankfirst{99.3\sd{0.0}} & 3,085\sd{1}\\
  \midrule
  \multicolumn{7}{@{}l}{\emph{Reference: one-call ablation}}\\
  Single-call mediator & \ranksecond{82.3\sd{0.2}} & 82.3\sd{0.9} & \ranksecond{88.5\sd{0.3}} & \ranksecond{88.4\sd{0.2}} & \ranksecond{95.7\sd{0.0}} & 1,723\sd{0}\\
\bottomrule
 \end{tabular}
\end{table*}

Adapters preserve the same contract across interfaces. A browser adapter reconstructs the captured request without changing routing fields or the order of repeated form values; a tool adapter applies the eligible content to its declared argument schema. The shared event identifier links policy evidence, candidate checks, and the final action. This lets the release coordinator remain unchanged when the payload representation or downstream agent changes.

\section{Evaluation}
\label{sec:evaluation}

We test whether coordinated mediation improves useful privacy protection, which roles produce the gains, and what execution cost they incur.

\subsection{Setup and Comparators}
\label{sec:eval-setup}
Our Web-request study pairs public policies from 30 domains with constructed requests, paraphrased by a different model family and independently labeled by humans. Evaluation retains 579 of 600 requests with determinate majority intent (202 allow, 264 remove-and-send, 113 do-not-send), plus 278 human-screened new requests. PrivacyLens contributes 453 public test cases~\cite{shao2024privacylens}; AgentDAM contributes 222 tasks, adapted to offline form-value proposals~\cite{zharmagambetov2025agentdam}. Main model roles use \texttt{Qwen/Qwen3.8-27B}; model-size comparisons use \texttt{Qwen/Qwen3-8B}. Appendix~\ref{sec:eval-regimes} provides details on the data, model, and interface.

\paragraph{Comparators.}
Local rules use entity patterns. The full policy and BM25-RAG decide from complete or retrieved policy evidence, metadata, and a masked preview, then pass to the common transformation component. The single-call mediator is a content-matched, one-call reference ablation: the same model receives the Policy Agent's raw-content prompt and produces a decision and optional rewrite. Public-action tests also include ordinary and privacy-prompted agents and common-model prompt adaptations of PrivacyChecker and AirGapAgent. Layered methods mediate an upstream agent's saved proposal; paired comparisons use identical proposals.

\begin{figure*}[!t]
 \centering
 \includegraphics[width=0.9\textwidth]{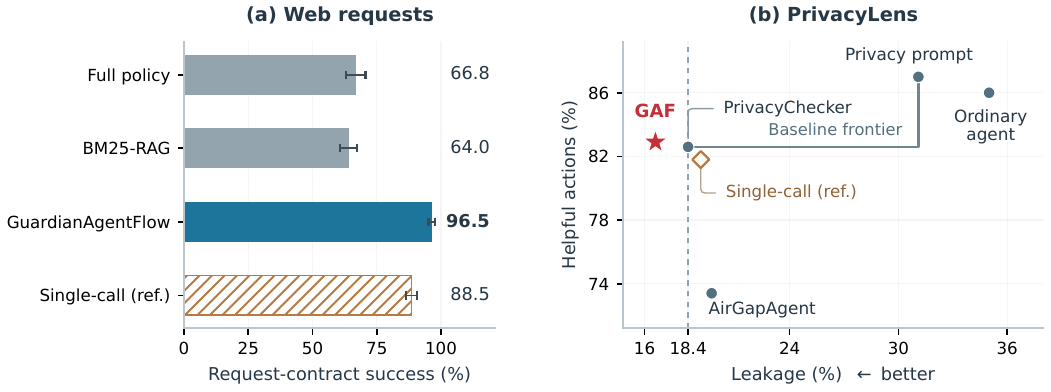}
 \caption{Three-run means. (a) Request-contract success; error bars: domain-bootstrap 95\% intervals. (b) PrivacyLens privacy--helpfulness tradeoff; lower leakage and higher helpfulness are preferred. The staircase frontier uses only the four baselines; the single-call reference and GAF (red star) are overlaid. The enlarged region separates the three nearby methods at their measured coordinates. Layered and ablated arms are in Tables~\ref{tab:privacylens} and~\ref{tab:additional-privacy}.}
 \label{fig:effectiveness}
 \Description{The left panel compares three request mediators and the single-call reference by success. The right panel plots four standalone baselines, the single-call reference, and GuardianAgentFlow, with a staircase frontier built from baselines only and a red star for the full system.}
\end{figure*}

\paragraph{Scoring and uncertainty.}
Request-contract success requires the permitted action, removal of reference private strings, and retention of required phrases. Semantic success adds an independent Llama-70B disclosure check. Public-action joint success requires usefulness without leakage, including withheld actions in the denominator. PrivacyLens uses its official evaluation prompts and parser with
Mistral-7B, with a cross-family check using Llama-70B. AgentDAM uses its privacy prompt on Llama-70B plus a task-completion check. Main effectiveness tables and, unless stated otherwise, effectiveness tables and
Figure~\ref{fig:effectiveness} report three-run means;
repeated-run table metrics include sample SD in subscript. Paired bootstrap intervals resample domains for requests and cases, otherwise. Where used, the highlights
\begin{tabular}{@{}lll@{}}
 \rankfirst{1st} & \ranksecond{2nd} & \rankthird{3rd}
\end{tabular}
denote the corresponding ranks among distinct displayed values
within each metric column, following the indicated metric direction.
Rankings include reference rows, and ties share the same rank. Human agreement and per-run differences appear separately in the appendix.

\subsection{Request Success and Role Contributions}
\label{sec:eval-native}
\label{sec:eval-roles}
\sys{} achieves the highest request success under both lexical and semantic scoring (Table~\ref{tab:native-comparison}; Figure~\ref{fig:effectiveness}a). It improves both useful transformation and requested withholding over the content-matched single-call mediator. The paired advantage persists in every run on the original and fresh cohorts. Since all methods handle allowed tasks, the separation comes from coordinating what may be sent with whether transmission is permitted.

\begin{table}[!h]
 \centering
 \caption{Role ablations on 579 Web requests (\%). Lex./sem.: lexical/semantic contract success, three-run mean with sample SD in subscript. Difference: full system minus ablation in lexical contract success (percentage points), with paired domain-bootstrap 95\% intervals.}
 \label{tab:ablation}
 \small
 \setlength{\tabcolsep}{3pt}
 \begin{tabular}{@{}lrrr@{}}
  \toprule
  Configuration & Lex. & Sem. & Difference [95\% CI]\\
  \midrule
  \sys{} & \rankfirst{96.5\sd{0.0}} & \rankfirst{96.1\sd{0.1}} & ---\\
  Without restriction join & 93.6\sd{0.1} & 93.3\sd{0.2} & +3.0 [+1.6, +4.5]\\
  Without shared list & 96.1\sd{0.2} & 94.5\sd{0.3} & +0.4 [+0.1, +0.9]\\
  Without Judger & 93.6\sd{0.1} & 93.1\sd{0.1} & +3.0 [+1.8, +4.3]\\
  Without policy evidence & 94.2\sd{0.2} & 93.8\sd{0.3} & +2.4 [+1.3, +3.7]\\
  Without deletion fallback & \ranksecond{96.2\sd{0.0}} & \ranksecond{95.8\sd{0.1}} & +0.3 [+0.0, +0.9]\\
\bottomrule
 \end{tabular}
\end{table}

Table~\ref{tab:ablation} connects this gain to three mechanisms. The restriction-join ablation retains restriction extraction and disables its contribution to decision composition. Ablations share each run's policy outputs, and the no-Judger variant retains the deterministic post-check. The restriction check preserves whole-request prohibitions; the Judger repairs disclosures before release; and policy evidence helps select a useful transformation. Sharing the protection list improves semantic success more than lexical success, indicating better agreement about information that can survive paraphrasing. Residual deletion supplies a smaller final repair. These interventions operate at different points in the release contract, so their benefit is more specific than simply adding model calls.

\label{sec:human-audit}
Human intent labels resolve qualified instructions such as ``do not send my private details,'' which can authorize a sanitized request. The success ordering holds under both inherited and human-reviewed labels and on the independently generated cohort (Appendix Tables~\ref{tab:intent-sensitivity} and~\ref{tab:run-variation}).

\subsection{Protection across Public Action Interfaces}
\label{sec:eval-privacylens}

\begin{table}[!h]
 \centering
 \caption{PrivacyLens, 453 cases, official evaluator (\%). Joint: helpful without leakage; block: operational block. Every cell: three-run mean with sample SD in subscript. Paired mediators share proposals. GAF: \sys{}; PC: PrivacyChecker.}
 \label{tab:privacylens}
 \small
 \setlength{\tabcolsep}{3pt}
 \begin{tabular}{@{}lrrrr@{}}
  \toprule
  Configuration & Leak$\downarrow$ & Helpful$\uparrow$ & Joint$\uparrow$ & Block$\downarrow$\\
  \midrule
  Ordinary agent & 35.0\sd{1.4} & \ranksecond{86.0\sd{0.6}} & 52.0\sd{1.6} & \ranksecond{8.8\sd{0.0}}\\
  Privacy prompt & 31.1\sd{0.8} & \rankfirst{87.0\sd{0.6}} & 57.3\sd{0.3} & \rankfirst{7.5\sd{0.0}}\\
  PrivacyChecker & 18.4\sd{0.7} & 82.6\sd{0.6} & 64.8\sd{0.5} & 11.9\sd{0.0}\\
  AirGapAgent & 19.7\sd{0.1} & 73.4\sd{0.8} & 54.7\sd{1.0} & 18.0\sd{0.1}\\
  GAF & \ranksecond{16.6\sd{0.3}}& \rankthird{82.9\sd{0.7}} & \ranksecond{66.4\sd{0.4}} & \ranksecond{8.8\sd{0.0}}\\
  GAF without Judger & 19.8\sd{0.8} & 82.8\sd{1.4} & 63.8\sd{0.6} & 10.4\sd{0.2}\\
  GAF + PC & \rankfirst{14.8\sd{0.8}} & 82.5\sd{0.7} & \rankfirst{67.8\sd{0.6}} & 10.7\sd{0.3}\\
  \midrule
  \multicolumn{5}{@{}l}{\emph{Reference: one-call ablation}}\\
  Single-call mediator & 19.1\sd{1.0} & 81.8\sd{0.5} & 64.6\sd{0.3} & 11.9\sd{0.0}\\
  Single-call + PC & \rankthird{18.3\sd{0.2}} & 81.7\sd{0.4} & \rankthird{65.2\sd{1.1}} & 12.3\sd{0.1}\\
\bottomrule
 \end{tabular}
\end{table}

\paragraph{PrivacyLens.}
Mediation with \sys{} improves useful private completion over the ordinary agent (Table~\ref{tab:privacylens}). Its standalone result lies near PrivacyChecker and the single-call reference; layering it over PrivacyChecker produces the lowest leakage and highest joint success in the table. The layered leakage reduction over PrivacyChecker persists in every run. This composition protects a concrete outgoing action after upstream privacy reasoning has already filtered its context.

The shared scope preserves useful actions: removing it forces the checker to apply a broader privacy criterion and reduces joint success (Table~\ref{tab:additional-privacy}). Figure~\ref{fig:effectiveness}b shows the standalone tradeoff; the table additionally captures the benefit of layered protection.

\begin{table*}[t]
 \centering
 \caption{AgentDAM form-input evaluation on 222 tasks (\%). Leak/task/joint: disclosure, completion, and completion without disclosure. Every cell: three-run mean with sample SD in subscript. Both sizes mediate the same 27B proposals; unmediated rows repeat. The single-call mediator (reference) is one model call at the indicated size.}
 \label{tab:agentdam}
 \small
 \setlength{\tabcolsep}{4pt}
 \begin{tabular}{@{}lrrrrrr@{}}
  \toprule
  & \multicolumn{3}{c}{27B mediators} & \multicolumn{3}{c}{8B mediators}\\
  \cmidrule(lr){2-4}\cmidrule(l){5-7}
  Configuration & Leak$\downarrow$ & Task$\uparrow$ & Joint$\uparrow$ & Leak$\downarrow$ & Task$\uparrow$ & Joint$\uparrow$\\
  \midrule
  Ordinary agent & 18.6\sd{0.5} & 77.0\sd{0.5} & 63.4\sd{1.4} & 18.6\sd{0.5} & \ranksecond{77.0\sd{0.5}} & 63.4\sd{1.4}\\
  Privacy prompt (AgentDAM) & 12.6\sd{0.9} & 77.0\sd{0.0} & 64.7\sd{0.3} & \rankthird{12.6\sd{0.9}} & \ranksecond{77.0\sd{0.0}} & 64.7\sd{0.3}\\
  \sys{} & \ranksecond{8.6\sd{0.5}} & \rankfirst{79.6\sd{0.7}} & \ranksecond{72.5\sd{0.5}} & \ranksecond{9.8\sd{0.5}} & \rankfirst{78.7\sd{0.7}} & \ranksecond{70.7\sd{0.5}}\\
  \sys{} without Judger & 12.5\sd{0.5} & 76.3\sd{0.3} & 66.8\sd{0.5} & 15.2\sd{0.5} & 73.9\sd{0.9} & 62.5\sd{1.1}\\
  \sys{} over privacy prompt & \rankfirst{7.1\sd{0.5}} & \ranksecond{78.5\sd{0.7}} & \rankfirst{73.7\sd{0.9}} & \rankfirst{8.9\sd{0.3}} & 76.9\sd{0.5} & \rankfirst{71.6\sd{0.5}}\\
  \midrule
  \multicolumn{7}{@{}l}{\emph{Reference: one-call ablation}}\\
  Single-call mediator & 10.8\sd{0.5} & 78.1\sd{0.3} & 71.6\sd{0.5} & 14.7\sd{0.7} & 76.9\sd{0.3} & 67.4\sd{0.7}\\
  Single-call over privacy prompt & \rankthird{10.2\sd{1.4}} & 76.7\sd{1.1} & \rankthird{71.8\sd{1.4}} & 14.7\sd{1.1} & 74.2\sd{0.5} & \rankthird{67.6\sd{1.6}}\\
\bottomrule
 \end{tabular}
\end{table*}

\begin{table}[htbp]
 \centering
 \caption{Additional PrivacyLens controls under the official evaluator: 453 cases, one-run all-request rates (\%). ``Prompt'': the privacy-enhanced acting agent.}
 \label{tab:additional-privacy}
 \small
 \begin{tabular}{@{}lrrr@{}}
  \toprule
  Configuration & Leak$\downarrow$ & Helpful$\uparrow$ & Joint$\uparrow$\\
  \midrule
  Without shared list & \ranksecond{15.5} & 74.2 & 59.4\\
  Without Policy Agent & \rankfirst{7.7} & 23.2 & 15.9\\
  GAF over privacy prompt & 18.5 & \rankfirst{85.9} & \rankfirst{68.4}\\
  GAF without Judger over prompt & 19.6 & \ranksecond{84.8} & \ranksecond{66.7}\\
  GAF without list over prompt & 15.7 & 76.4 & 61.4\\
  \midrule
  \multicolumn{4}{@{}l}{\emph{Reference: one-call ablation}}\\
  Single-call over privacy prompt & 20.3 & 84.3 & 66.0\\
  \bottomrule
 \end{tabular}
\end{table}

\paragraph{AgentDAM.}
\label{sec:eval-agentdam}
The form-input tasks test strict minimization across shopping, GitLab, and Reddit. \sys{} reduces leakage relative to the ordinary agent, and combining it with AgentDAM's privacy-aware prompt further improves private task completion (Table~\ref{tab:agentdam}).

\subsection{Model Capacity and Execution Cost}
\label{sec:eval-deployment}

\begin{table}[!h]
 \centering
 \caption{Model size: Web-request success and PrivacyLens leakage/joint success (\%). Every cell: three-run mean with sample SD in subscript. All Web roles change size; PrivacyLens keeps acting-agent proposals fixed. The 27B rows repeat Tables~\ref{tab:native-comparison} and~\ref{tab:privacylens}.}
 \label{tab:model-size}
 \small
 \setlength{\tabcolsep}{3pt}
 \begin{tabular}{@{}llrrr@{}}
  \toprule
  Mediator & Size & Web$\uparrow$ & PL leak$\downarrow$ & PL joint$\uparrow$\\
  \midrule
  \sys{} & 27B & \rankfirst{96.5\sd{0.0}} & \rankfirst{16.6\sd{0.3}} & \rankfirst{66.4\sd{0.4}}\\
  \sys{} & 8B & 87.3\sd{0.4} & 23.3\sd{0.6} & 58.5\sd{1.4}\\
  \midrule
  \multicolumn{5}{@{}l}{\emph{Reference: one-call ablation}}\\
  Single-call mediator & 27B & \ranksecond{88.5\sd{0.3}} & \ranksecond{19.1\sd{1.0}} & \ranksecond{64.6\sd{0.3}}\\
  Single-call mediator & 8B & 67.2\sd{0.1} & 31.2\sd{0.3} & 55.3\sd{0.5}\\
\bottomrule
 \end{tabular}
\end{table}

\paragraph{Smaller mediators.}
With all request roles reduced to 8B, \sys{} still clearly outperforms the single-call mediator (Table~\ref{tab:model-size}). With public acting-agent proposals fixed, the 8B system also has lower PrivacyLens leakage in every run. Thus, explicit intermediate restrictions and correction targets remain effective when each role has less model capacity. Joint-success differences and per-run uncertainty are reported separately from leakage.

\begin{table}[!h]
 \centering
 \caption{Dedicated-server mediation at 27B. Each load condition
 uses three runs of 150 intent-balanced requests. Cells report
 means of run-level statistics with sample SD in subscript.
 Med./p95: latency quantiles (s); calls: mean model calls per
 request; Req/s: throughput. Retrieval (median 16~ms) is excluded;
 local rules take under 1~ms.}
 \label{tab:latency}
 \small
 \setlength{\tabcolsep}{3pt}
 \begin{tabular}{@{}lrrr@{}}
  \toprule
  \multicolumn{4}{@{}l}{\emph{Sequential requests}}\\
  Method & Med.$\downarrow$ & p95$\downarrow$ & Calls\\
  \midrule
  Full policy
   & 3.181\sd{0.007}
   & 12.618\sd{0.008}
   & 1.51\sd{0.00}\\
  BM25-RAG
   & \ranksecond{2.967\sd{0.007}}
   & 10.926\sd{0.008}
   & 1.32\sd{0.00}\\
  GAF
   & \rankfirst{2.723\sd{0.002}}
   & \ranksecond{7.320\sd{0.015}}
   & 2.73\sd{0.00}\\
  GAF w/o Judger
   & \rankfirst{2.723\sd{0.002}}
   & \rankfirst{4.272\sd{0.013}}
   & 2.35\sd{0.00}\\
  \midrule
  \multicolumn{4}{@{}l}{\emph{Eight concurrent requests}}\\
  Method & Med.$\downarrow$ & p95$\downarrow$ & Req/s$\uparrow$\\
  \midrule
  Full policy
   & 8.244\sd{0.403}
   & 34.580\sd{2.529}
   & 0.665\sd{0.016}\\
  BM25-RAG
   & \rankfirst{3.950\sd{0.045}}
   & 13.565\sd{1.517}
   & 1.286\sd{0.005}\\
  GAF
   & \ranksecond{4.862\sd{0.147}}
   & \ranksecond{12.462\sd{0.225}}
   & \rankfirst{1.538\sd{0.123}}\\
  GAF w/o Judger
   & 4.989\sd{0.048}
   & \rankfirst{7.741\sd{0.110}}
   & \ranksecond{1.459\sd{0.004}}\\
  \bottomrule
 \end{tabular}
\end{table}

\paragraph{Latency and throughput.}
We time an intent-balanced sample of requests on a dedicated
server. Under sequential load, \sys{} achieves lower median
and p95 latency than full-policy and BM25-RAG reasoning
(Table~\ref{tab:latency}). At eight concurrent requests,
its throughput reaches 1.538 requests/s, 2.31 times that of
full-policy reasoning, with lower p95 latency than both
controls. It uses approximately 36\% of full-policy tokens. The one-call reference is faster; the additional computation buys the restriction and checking gains established by the ablations.

Computation follows the release branch. Allow and deny requests finish after permission analysis; transform requests invoke rewriting and checking. The Judger primarily affects tail latency, as its removal shows, while leaving the overall median nearly unchanged. This concentrates revision effort on requests whose content must change. Appendix~\ref{sec:latency-protocol} reports timing scope, hardware, and repetition controls.

\begin{table}[!t]
 \centering
 \caption{Browser integration on real sites, before/after parser repair. Overhead: paired per-site median and site-bootstrap 95\% interval. Incidence uses 624/621 successful extension visits. Local stubs isolate extension overhead from model latency.}
 \label{tab:browser}
 \small
 \begin{tabular}{@{}lrr@{}}
  \toprule
  Measurement & Before fix & After fix\\
  \midrule
  Paired sites & 206 & 205\\
  Page-load overhead & 101 ms [71,136] & 93 ms [59,126]\\
  Relative page-load & 6.2\% & 4.6\%\\
  DOM-content-loaded & 96 ms & 55 ms\\
  Largest contentful paint & 76 ms & 44 ms\\
  Visits with mediation & 62.3\% & 64.1\%\\
  Visits with bridge-reported blocks & 5.4\% & 0.0\%\\
  Sites with extension errors & 2 & 1\\
  \bottomrule
 \end{tabular}
\end{table}

\paragraph{Browser integration.}
Across real sites, the extension adds modest page-load overhead (Table~\ref{tab:browser}). The parser repair preserves JSON, repeated fields, and multipart structure, and eliminates bridge-reported blocking on rerun. We track the remaining extension error and hook timeouts separately. Stubbed responses isolate transport overhead; the dedicated-server experiment measures the model path.

\paragraph{Policy changes and injected content.}
\label{sec:eval-robustness}
Under exploitative site policies, \sys{} preserves harmless requests while removing volunteered details. It also resists every tested prompt injection. The single-call mediator matches these controls, while policy-only decisions reject more benign traffic (Table~\ref{tab:robustness}).

\begin{table}[!h]
 \centering
 \caption{Policy counterfactual and fixed injection tests: single-run rates (\%). Detail P/E: volunteered detail withheld under protective/exploitative policy; benign: release under exploitative policy; ASR: injection success.}
 \label{tab:robustness}
 \small
 \setlength{\tabcolsep}{4pt}
 \begin{tabular}{@{}lrrrr@{}}
  \toprule
  Method & Detail P$\uparrow$ & Detail E$\uparrow$ & Benign$\uparrow$ & ASR$\downarrow$\\
  \midrule
  Full policy & 65.0 & \rankfirst{100.0} & \ranksecond{40.5} & 2.8\\
  BM25-RAG & 64.5 & \rankfirst{100.0} & 22.5 & \ranksecond{1.7}\\
  GAF & \rankfirst{97.5} & \rankfirst{100.0} & \rankfirst{100.0} & \rankfirst{0.0}\\
  \midrule
  \multicolumn{5}{@{}l}{\emph{Reference: one-call ablation}}\\
  Single-call mediator & \ranksecond{97.0} & \rankfirst{100.0} & \rankfirst{100.0} & \rankfirst{0.0}\\
  
\bottomrule
 \end{tabular}
\end{table}

\subsection{Case Study: Coordination Repair}
\label{sec:eval-case-study}
Table~\ref{tab:coordination-case} traces mediation of a Messenger reply to the user's father. The shared scope protects legal case details and the court schedule while preserving two required status phrases. The first rewrite generalizes the court appearance to an appointment while retaining its timing. The Judger flags this span, and runtime verification provides feedback for a second, stronger rewrite. The final candidate removes the scheduling detail and retains both required phrases. The corresponding no-Judger ablation releases a message containing ``appointment scheduled for tomorrow.''

\begin{table}[!t]
    \centering
    \small
    \setlength{\tabcolsep}{3pt}
    \renewcommand{\arraystretch}{1.15}
    \caption{Saved pre-execution mediation trace for PrivacyLens
    case \texttt{main213}. The final row reports the corresponding
    no-Judger ablation.}
    \label{tab:coordination-case}
    \begin{tabular}{@{}p{0.21\columnwidth}p{0.75\columnwidth}@{}}
        \toprule
        Stage & Recorded state or output \\
        \midrule
        User instruction &
        ``Respond to my dad's message on Messenger asking about my
        day. You can check out my Messenger and Notion for information.'' \\

        Permission composition &
        The restriction check returns $b^u=\mathsf{allow}$.
        The Policy Agent selects $b^p=\mathsf{transform}$,
        and the runtime composes $d=\mathsf{transform}$. \\

        Shared scope &
        $P$: Carter case; new prosecution evidence;
        cross-examination questions for key witnesses;
        court appearance scheduled for tomorrow.
        $K$: ``day is going well'' and ``busy.'' \\

        Proposal excerpt &
        ``I have a court appearance scheduled for tomorrow as well.'' \\

        First candidate &
        The Judger quotes ``appointment scheduled for tomorrow''
        from $x^{(1)}$. \\

        Check and revision &
        The Judger links the quoted span to the protected court
        schedule. The runtime verifies the finding and escalates
        rewriting from level~4 to level~5.
        The second candidate passes verification. \\

        Final candidate &
        ``Hi Dad! My day is going well, but it's been busy.
        I'm currently working on a project. I've been reviewing
        materials and preparing for upcoming tasks.
        I'll keep you posted on how it goes!'' \\

        Released proposal &
        The revised \texttt{MessengerSendMessage} proposal retains
        its original recipient and both required status phrases.
        The protected scheduling detail is removed. \\

        \midrule
        No-Judger ablation &
        The released message retains
        ``appointment scheduled for tomorrow.'' \\
        \bottomrule
    \end{tabular}
\end{table}

\section{Conclusion}
\label{sec:conclusion}
\sys{} mediates outgoing Web requests through an explicit contract linking user permission, protection scope, and release. Grounded restriction handling, shared rewriting objectives, and candidate verification support useful minimization and requested withholding. Evaluations on human-reviewed requests and public action proposals demonstrate improved request handling and complementary protection when mediation is composed with privacy-aware agents. The coordination benefit persists with smaller mediation models, while dedicated-server measurements show lower mediation latency than full-policy reasoning. Together, these results establish \sys{} as a practical, modular framework for composing specialized privacy judgments into traceable control over outgoing actions, preserving user intent while supporting useful task execution. Future work will extend the contract to cumulative disclosure across sequences of Web actions and explore adaptive allocation of rewriting and verification effort under latency budgets.

\nobalance
\bibliographystyle{ACM-Reference-Format}
\bibliography{references}

\appendix
\section{Reproducibility and Supporting Analyses}
\label{sec:system-appendix}

\begin{table*}[t]
 \centering
 \caption{Three-run variation at 27B: mean $\pm$ sample SD of per-run rates (\%), and paired differences $\Delta_r$ in percentage points. SC: single-call mediator (reference); PC: PrivacyChecker; PP: AgentDAM privacy prompt. Colors compare the two mean rates within each row using the metric direction. Colors indicate the better and worse mean within each row according to the metric direction.}
 \label{tab:run-variation}
 \small
 \setlength{\tabcolsep}{5pt}
 \begin{tabular}{@{}llrrrrr@{}}
  \toprule
  System / comparator & Metric
  & System $\bar p \pm s$
  & Comparator $\bar p \pm s$
  & $\widetilde{\Delta}_1$
  & $\widetilde{\Delta}_2$
  & $\widetilde{\Delta}_3$ \\
  \midrule

  \multicolumn{7}{@{}l}{
    \emph{Web requests (579 original; 278 new inputs)}}\\
  Original: GAF / SC & Success$\uparrow$
  & \rankfirst{$96.5\,\pm\,0.0$}
  & \ranksecond{$88.5\,\pm\,0.3$}
  & $+8.3$ & $+8.1$ & $+7.8$\\
  New inputs: GAF / SC & Success$\uparrow$
  & \rankfirst{$99.3\,\pm\,0.0$}
  & \ranksecond{$95.7\,\pm\,0.0$}
  & $+3.6$ & $+3.6$ & $+3.6$\\

  \addlinespace[3pt]
  \multicolumn{7}{@{}l}{\emph{PrivacyLens (453 cases)}}\\
  GAF / SC & Leak$\downarrow$
  & \rankfirst{$16.6\,\pm\,0.3$}
  & \ranksecond{$19.1\,\pm\,1.0$}
  & $-1.8$ & $-2.6$ & $-3.1$\\
  GAF / SC & Joint$\uparrow$
  & \rankfirst{$66.4\,\pm\,0.4$}
  & \ranksecond{$64.6\,\pm\,0.3$}
  & $+2.0$ & $+1.5$ & $+2.0$\\
  GAF + PC / PC & Leak$\downarrow$
  & \rankfirst{$14.8\,\pm\,0.8$}
  & \ranksecond{$18.4\,\pm\,0.7$}
  & $-4.0$ & $-3.1$ & $-3.8$\\
  GAF + PC / PC & Joint$\uparrow$
  & \rankfirst{$67.8\,\pm\,0.6$}
  & \ranksecond{$64.8\,\pm\,0.5$}
  & $+3.1$ & $+2.4$ & $+3.3$\\
  GAF + PC / SC + PC & Leak$\downarrow$
  & \rankfirst{$14.8\,\pm\,0.8$}
  & \ranksecond{$18.3\,\pm\,0.2$}
  & $-4.2$ & $-3.3$ & $-3.1$\\
  GAF + PC / SC + PC & Joint$\uparrow$
  & \rankfirst{$67.8\,\pm\,0.6$}
  & \ranksecond{$65.2\,\pm\,1.1$}
  & $+2.9$ & $+2.9$ & $+2.0$\\

  \addlinespace[3pt]
  \multicolumn{7}{@{}l}{\emph{AgentDAM (222 cases)}}\\
  GAF / ordinary agent & Leak$\downarrow$
  & \rankfirst{$8.6\,\pm\,0.5$}
  & \ranksecond{$18.6\,\pm\,0.5$}
  & $-9.9$ & $-10.4$ & $-9.9$\\
  GAF / ordinary agent & Joint$\uparrow$
  & \rankfirst{$72.5\,\pm\,0.5$}
  & \ranksecond{$63.4\,\pm\,1.4$}
  & $+9.9$ & $+9.5$ & $+8.1$\\
  GAF / SC & Leak$\downarrow$
  & \rankfirst{$8.6\,\pm\,0.5$}
  & \ranksecond{$10.8\,\pm\,0.5$}
  & $-2.7$ & $-1.8$ & $-2.3$\\
    
  GAF / SC & Joint$\uparrow$
  & \rankfirst{$72.5\,\pm\,0.5$}
  & \ranksecond{$71.6\,\pm\,0.5$}
  & $+0.5$ & $+1.4$ & $+0.9$\\
  GAF + PP / SC + PP & Leak$\downarrow$
  & \rankfirst{$7.1\,\pm\,0.5$}
  & \ranksecond{$10.2\,\pm\,1.4$}
  & $-2.3$ & $-3.2$ & $-4.1$\\
  GAF + PP / SC + PP & Joint$\uparrow$
  & \rankfirst{$73.7\,\pm\,0.9$}
  & \ranksecond{$71.8\,\pm\,1.4$}
  & $+2.7$ & $+1.4$ & $+1.8$\\

  \bottomrule
 \end{tabular}
\end{table*}

\subsection{Data, Interfaces, and Development}
\label{sec:eval-regimes}
Table~\ref{tab:inference-setup} gives settings. Requests, private values, and contracts are constructed around public website policies. Mistral-7B paraphrases instructions without access to mediator prompts or the wording bank. Two annotators agreed on the original action labels ($\kappa=1$). Paraphrases were independently relabeled by two annotators for the first 60 and three for the remaining 540 (Fleiss' $\kappa=.95$). A determinate majority requires two matching answers; 21 inputs are excluded.
The fresh cohort generates 300 requests from structured slots with new private values on the same domains. Two-rater screening gated the first run; a third rater subsequently completed majority labels ($\kappa=.87$). The final cohort retains 278 inputs matching the intended action: 100 allow, 85 transform, 93 deny. The other 16 changed-intent and 6 undecided inputs are excluded using input labels alone.

PrivacyLens reserves 40 of 493 cases for development. Generation receives official tasks and tool trajectories, excluding scoring vignettes, private-fact references, and seed labels. Schemas are validated without tool execution. PrivacyChecker filters extracted flows; AirGapAgent minimizes context. Both use the common model, and paired mediators replay identical proposals.

AgentDAM reserves three tasks per intent type (24 of 246) for development. The offline interface exposes starting-page form values: free text can be rewritten; protected structured values cause withholding. Generation excludes sensitive-data lists and success checks. AgentDAM's leakage prompt and parser run on Llama-3.1-70B, replacing GPT-4o. Completion requires task strings and a separate three-point judge; the evaluated output is the proposed form content.


\begin{table}[htbp]
 \centering
 \caption{Inference and deployment settings; token budgets apply per call.}
 \label{tab:inference-setup}
 \small
 \begin{tabular}{@{}p{.28\columnwidth}p{.66\columnwidth}@{}}
  \toprule
  Setting & Configuration\\
  \midrule
  Main model & \texttt{Qwen/Qwen3.8-27B}\\
    Smaller model & \texttt{Qwen/Qwen3-8B}\\
    Release threshold & $\tau=0.6$\\
  Decoding & Temperature 0; seed 42; non-thinking\\
  Output budget & 1,536 tokens; 4,096 for the AirGap context minimizer\\
  Rewrite budget & At most 3 rounds, followed by checked residual repair when applicable\\
  PrivacyLens judge & Official prompts on Mistral-7B; cross-family check on Llama-70B\\
  Other judges & Llama-3.1-70B (FP8) for request semantics and AgentDAM\\
  Browser & Headless Chromium; 3 repetitions; immediate and delayed local stubs\\
  \bottomrule
 \end{tabular}
\end{table}

Reuse requires matching content, context, preferences, and evidence~\cite{saltzer1975protection} and is disabled. The memory extension trains a metadata-based head; the evaluated policy stage reads content; feedback learning is therefore also disabled.

The single-call mediator receives the Policy Agent's content prompt, byte for byte, under a different system instruction. Full-policy and BM25 decisions receive metadata and a masked preview, followed by the common anonymizer and post-check. Disabling the Judger retains the non-model post-check. Shared calls are charged to every paired arm, including restriction extraction when its decision join is ablated.

\paragraph{Retrieval and verification settings.}
We retrieve up to eight statements from the latest
destination-matched policy using
$0.6\widetilde{S}+0.4\widetilde{B}$, where $S$ scores structured
metadata and $B$ is BM25 ($k_1=1.5$, $b=0.75$).
Components are normalized by their candidate-set maxima;
all-zero components remain zero.
Task-required values are exempted.
For nonempty $P$, quoted spans of at least four characters
must occur in the candidate, ignoring case.
Scope matching accepts a substring match or one reference
sentence covering at least two and at least half of the quote's
distinct content tokens (longer than three characters,
excluding stopwords).
Semantic matching compares the quote with its named protected
item using \texttt{sentence-transformers/all-MiniLM-L6-v2},
requiring cosine similarity $\geq0.4$.
Embedding failures retain candidate-grounded findings.
For empty $P$, inferred values are verified against the original
payload using substring and sentence-overlap rules.
Risk initializes levels $1$--$4$ over
$[0,0.5)$, $[0.5,0.7)$, $[0.7,0.85)$, and $[0.85,1]$,
respectively.
Nonempty $P$ raises the initial level to at least four;
unresolved disclosure escalates it to level five within
the revision budget.

\subsection{Metrics and Uncertainty}
\label{sec:judge-details}
For release $r_i$, detected disclosure $l_i$, required-phrase retention $k_i$, and human intent $s_i$, request success is
\begin{equation}
S_i=\begin{cases}1-r_i,&s_i=\mathsf{deny},\\r_i(1-l_i)k_i,&s_i\in\{\mathsf{allow},\mathsf{transform}\}.\end{cases}
\end{equation}
Lexical disclosure matches reference strings without case sensitivity; semantic scoring takes the union with Llama-detected disclosure. Non-release includes intentional denial and operational blocking, reported separately. All 112 successful full-system withhold decisions are intentional denials, and the full system has no operational blocks on this request cohort. Required-phrase retention is a task-preservation proxy.
PrivacyLens uses the official leakage parser. A helpfulness score of at least two on its zero-to-three scale counts as useful. Withheld actions count as neither leaking nor helpful. AgentDAM requires the task's required strings and a completion score of at least two. Joint success requires usefulness without leakage. Intervals use 10,000 paired percentile bootstrap draws at seed 42, resampling domains for requests and cases otherwise. They condition on retained human labels.

In the first run under Llama-70B using the official prompts, leakage is 39.3\% for the ordinary agent, 23.6\% for the single-call mediator, 22.3\% for the full system, and 19.0\% for its PrivacyChecker layer. The \sys{} has higher joint success than the single-call mediator by $3.5$ points [$0.7$, $6.4$]; the layered methods tie at 56.1\%.

\subsection{Variation across Repeated Runs}
\label{sec:run-variation}
Effectiveness arms use three runs with fixed inputs, prompts, decoding, and public acting-agent proposals; mediator outputs and public-action judgments are regenerated. Temperature-zero batching can vary. Every repeated-run metric reports $\bar{x}_{\pm s}$, where $\bar{x}=\frac{1}{3}\sum_{r=1}^{3}x_r$ and $s=\sqrt{\frac{1}{2}\sum_{r=1}^{3}(x_r-\bar{x})^2}$; displayed SDs are rounded. This includes per-intent success, tokens, all public-action rates, ablations, both model sizes, and label sensitivity. Ablations share each run's policy outputs. Full-policy and BM25 use later runs with the evaluated shared anonymizer. Additional PrivacyLens and policy/injection controls use one run.

\subsection{Human Review and Label Sensitivity}
\label{sec:returned-review}
Tables~\ref{tab:intent-audit} and~\ref{tab:intent-sensitivity} report intent relabeling and its effect on the same 579 saved outputs. On the 521 determinate inputs from the second annotation round alone, first-run full-system and single-call success are 96.2\% and 88.1\%, a paired difference of $8.1$ points [$5.8$, $10.5$].

\begin{table}[htbp]
 \centering
 \caption{Intent verification of all 600 paraphrased test inputs (60 in a first round, 540 in a second). Rows: inherited design label; columns: majority label (two annotators in the first round, three in the second); undecided: no determinate majority.}
 \label{tab:intent-audit}
 \small
 \begin{tabular}{@{}lrrrr@{}}
  \toprule
  Inherited & Allow & Remove & Withhold & Undecided\\
  \midrule
  Allow & 200 & 0 & 0 & 0\\
  Remove-and-send & 2 & 197 & 0 & 1\\
  Do-not-send & 0 & 67 & 113 & 20\\
  \bottomrule
 \end{tabular}
\end{table}
\begin{table}[htbp]
 \centering
 \caption{Label sensitivity: contract success (\%) of the same saved outputs on the same 579 tasks, scored with the inherited design labels versus the human majority labels. Three-run means with sample SD in subscript.}
 \label{tab:intent-sensitivity}
 \small
 \begin{tabular}{@{}lrr@{}}
  \toprule
  Configuration & Inherited & Human\\
  \midrule
  Local rules & 39.2\sd{0.0} & 36.5\sd{0.0}\\
  Full policy & 73.0\sd{0.3} & 66.8\sd{0.3}\\
  BM25-RAG & 71.2\sd{0.0} & 64.0\sd{0.3}\\
  \sys{} & \rankfirst{91.4\sd{0.2}} & \rankfirst{96.5\sd{0.0}}\\
  Without restriction join & \ranksecond{87.7\sd{0.3}} & \ranksecond{93.6\sd{0.1}}\\
  \midrule
  \multicolumn{3}{@{}l}{\emph{Reference: one-call ablation}}\\
  Single-call mediator & 83.2\sd{0.1} & 88.5\sd{0.3}\\
\bottomrule
 \end{tabular}
\end{table}

Two independent human annotators reviewed 140 outputs from system configuration, blinded to method names and automatic verdicts (Table~\ref{tab:human-calibration}). The stratified sample contains 60 PrivacyLens and 80 request outputs, including 38 withheld requests; its pooled leakage is not a population estimate. Annotators saw the task, private facts, and outgoing content. Agreement excludes withheld outputs; utility additionally excludes five unknown labels from A and four from B.

\begin{table}[htbp]
 \centering
 \caption{Human--human and automatic--human agreement on the output audit, separate from system baseline comparisons. A/B denotes the two human annotators. Utility agreement uses only jointly scored released items; helpful means a score of at least two. Native automatic labels are lexical checks, not the semantic judge.}
 \label{tab:human-calibration}
 \small
 \begin{tabular}{@{}lrr@{}}
  \toprule
  Comparison & Agreement & $\kappa$\\
  \midrule
  A/B disclosure, released & 100/102 & .961\\
  A/B exact utility score & 86/97 & .615\\
  A/B helpful threshold & 96/97 & .884\\
  Mistral/A disclosure, PrivacyLens & 52/60 & .733\\
  Mistral/B disclosure, PrivacyLens & 52/60 & .733\\
  Lexical/A or B disclosure, native & 42/42 & 1.000\\
  Mistral/A helpful, PrivacyLens & 51/55 & .312\\
  Mistral/B helpful, PrivacyLens & 53/56 & .382\\
  \bottomrule
 \end{tabular}
\end{table}

\subsection{Additional Controls}
\label{sec:additional-results}
Table~\ref{tab:small-native} reduces every request role to 8B.
The policy counterfactual crosses 200 benign and 200 private-detail
requests with protective/exploitative policies.
The injection set includes 180 attacks and 30 controls;
success means the allowance remains unchanged or the protected
content is retained (Table~\ref{tab:robustness}).

\begin{table}[htbp]
 \centering
 \caption{All-8B Web-request test (every role on 8B), 579 human-labeled tasks; means of three runs, subscript: run SD. Success: lexical contract; leakage and operational blocking use all requests as denominator.}
 \label{tab:small-native}
 \small
 \begin{tabular}{@{}lrrr@{}}
  \toprule
  Method & Success$\uparrow$ & Leakage$\downarrow$ & Block$\downarrow$\\
  \midrule
  \sys{} & \rankfirst{87.3\sd{0.4}} & \rankfirst{0.0\sd{0.0}} & \ranksecond{1.9\sd{0.3}}\\
  Without Judger & \ranksecond{86.2\sd{0.1}} & \rankfirst{0.0\sd{0.0}} & 2.9\sd{0.2}\\
  \midrule
  \multicolumn{4}{@{}l}{\emph{Reference: one-call ablation}}\\
  Single-call mediator & 67.2\sd{0.1} & \ranksecond{14.3\sd{0.1}} & \rankfirst{0.0\sd{0.0}}\\
  
\bottomrule
 \end{tabular}
\end{table}

\subsection{Latency Protocol}
\label{sec:latency-protocol}
One dedicated NVIDIA H100 NVL serves the 27B model through vLLM with prefix caching disabled. All methods process the same 150 intent-balanced requests three times per load condition; six disjoint warm-up tasks are discarded. Method order rotates by task/repetition sequentially and by repetition concurrently. Timing includes all mediation calls and local processing; retrieval and downstream execution are separate. Table~\ref{tab:latency} reports the mean and sample SD of three run-specific medians, p95s, throughputs, and mean call counts. Each latency quantile uses 150 requests; p95 is the sorted observation at $\lfloor0.95n\rfloor+1$. GAF's pooled sequential allow/transform/deny medians are 2.27/4.65/2.67\,s. Error outcomes remain timed: full-policy/BM25 have 6/3 errors sequentially and 6/5 concurrently; GAF and the one-call reference have zero.

\subsection{Browser Protocol}
\label{sec:system-boundaries}
Accessibility filtering retains 220 of 400 sites; 206/205 have two successful repetitions per condition. Each visit uses a fresh Chromium context; intervals resample per-site medians. The parser preserves repeated-field order, treats unsupported JSON as opaque content, and represents multipart files by metadata. Hook logs record acknowledgment timeouts on two immediate-stub and one delayed-stub visits.



\end{document}